\documentclass[11pt,a4paper]{article}
\usepackage[hyperref]{acl2021}
\usepackage{times}
\usepackage{latexsym}
\usepackage{multirow}
\usepackage{graphicx}

\usepackage{booktabs}
\usepackage{xcolor}
\usepackage{setspace}

\usepackage{array}
\newcommand{\PreserveBackslash}[1]{\let\temp=\\#1\let\\=\temp}
\newcolumntype{C}[1]{>{\PreserveBackslash\centering}p{#1}}

\usepackage{amsmath,amssymb}

\allowdisplaybreaks

\usepackage{microtype}

\aclfinalcopy 
\def\aclpaperid{3057} 

\newcommand\blfootnote[1]{%
  \begingroup
  \renewcommand\thefootnote{}\footnote{#1}%
  \addtocounter{footnote}{-1}%
  \endgroup
}

\title{BERT is to NLP what AlexNet is to CV: \\ Can Pre-Trained Language Models Identify Analogies?}
\author{Asahi Ushio, Luis Espinosa-Anke, Steven Schockaert, Jose Camacho-Collados \\
  Cardiff NLP, School of Computer Science and Informatics \\
  Cardiff University, United Kingdom \\
  \normalsize \texttt{\{UshioA,Espinosa-AnkeL,SchockaertS1,CamachoColladosJ\}@cardiff.ac.uk}
  \\}
\date{}

\begin{document}

\maketitle


\blfootnote{While the title is probably self-explanatory, this is a small note explaining it. \textit{BERT is to NLP what AlexNet is to CV} is making an analogy on what the BERT and AlexNet models represented for Natural Language Processing (NLP) and Computer Vision (CV), respectively. They both brought a paradigm shift in how research was undertaken in their corresponding disciplines and this is what the analogy refers to.}

\begin{abstract}

Analogies play a central role in human commonsense reasoning. The ability to recognize analogies such as “eye is to seeing what ear is to hearing”, sometimes referred to as analogical proportions, shape how we structure knowledge and understand language.  Surprisingly, however, the task of identifying such analogies has not yet received much attention in the language model era. In this paper, we analyze the capabilities of transformer-based language models on this unsupervised task, using benchmarks obtained from educational settings, as well as more commonly used datasets. We find that off-the-shelf language models can identify analogies to a certain extent, but struggle with abstract and complex relations, and results are highly sensitive to model architecture and hyperparameters. Overall the best results were obtained with GPT-2 and RoBERTa, while configurations using BERT were not able to outperform word embedding models. Our results raise important questions for future work about how, and to what extent, pre-trained language models capture knowledge about abstract semantic relations.\footnote{Source code and data to reproduce our experimental results are available in the following repository:  \url{https://github.com/asahi417/analogy-language-model}}
\end{abstract}

\section{Introduction}

\begin{table}[t]
\centering
\footnotesize
\scalebox{1.05}{
\begin{tabular}{lcl}
\toprule
Query:       && word:language \\
\midrule
Candidates: & (1) & paint:portrait \\
 & (2) & poetry:rhythm \\
 & {\bf (3)} & {\bf note:music} \\
 & (4) & tale:story \\
 & (5) & week:year \\
\bottomrule
\end{tabular}
}
\caption{\label{tab:sat-sample}
An example analogy task from the SAT dataset. The third candidate is the answer to the query. }
\end{table}

One of the most widely discussed properties of word embeddings has been their surprising ability to model certain types of relational similarities in terms of word vector differences \cite{mikolov2013distributed,vylomova2016take,allen2019analogies,ethayarajh2019towards}. The underlying assumption is that when ``$a$ is to $b$ what $c$ is to $d$'' the word vector differences $\mathbf{b}-\mathbf{a}$ and $\mathbf{d}-\mathbf{c}$ are expected to be similar, where we write $\mathbf{x}$ for the embedding of a word $x$. While this assumption holds for some types of syntactic relations, for semantic relations this holds to a much more limited degree than was suggested in early work \cite{linzen2016issues,schluter2018word}. Moreover, the most commonly used benchmarks have focused on specific and well-defined semantic relations such as ``capital of'', rather than the more abstract notion of relational similarity that is often needed for solving the kind of psychometric analogy problems that can be found in IQ tests and educational settings. An example of such a problem is shown in Table \ref{tab:sat-sample}.

Given the central role of analogy in human cognition, it is nonetheless important to understand the extent to which NLP models are able to solve these more abstract analogy problems. Besides its value as an intrinsic benchmark for lexical semantics, the ability to recognize analogies is indeed important in the contexts of human creativity \cite{holyoak1996mental}, innovation \cite{hope2017accelerating}, computational creativity \cite{goel2019computational} and education \cite{pardos2020universitymap}. Analogies are also a prerequisite to build AI systems for the legal domain \cite{ashley1988arguing,walton2010similarity} and are used in machine learning \cite{miclet2008analogical,hug2016analogical,hullermeier2020towards} and for ontology alignment \cite{raad2015role}, among  others.

Within NLP, however, the task of recognizing analogies has received relatively little attention. To solve such problems, \citet{Turney:2005:MSS:1642293.1642475} proposed Latent Relational Analysis (LRA), which was essentially designed as a relational counterpart to Latent Semantic Analysis \cite{Landauer:1997}. 
Somewhat surprisingly, perhaps, despite the substantial progress that word embeddings and language models (LMs) have enabled in NLP, LRA still represents the current state-of-the-art in solving abstract word analogy problems. When going beyond a purely unsupervised setting, however, GPT-3 was recently found to obtain slightly better results \cite{DBLP:conf/nips/BrownMRSKDNSSAA20}.

The aim of this paper is to analyze the ability of pre-trained LMs to recognize analogies. Our focus is on the zero-shot setting, where LMs are used without fine-tuning. To predict whether two word pairs $(a, b)$ and $(c ,d)$ are likely to be analogical, we need a prompt, i.e.\ a template that is used to construct the input to the LM, and a scoring function. We extensively analyze the impact of both of these choices, as well as the differences between different LMs. When the prompt and scoring function are carefully calibrated, we find that 
GPT-2 can outperform LRA, standard word embeddings as well as the published results for GPT-3 in the zero-shot setting.
However, we also find that these results are highly sensitive to the choice of the prompt, as well as two hyperparameters in our scoring function, with the optimal choices not being consistent across different datasets. Moreover, using BERT leads to considerably weaker results, underperforming even standard word embeddings in all of the considered configurations.
These findings suggest that while transformer-based LMs learn relational knowledge to a meaningful extent, more work is needed to understand how such knowledge is encoded, and how it can be exploited.

\section{Related work}

\subsection{Understanding Pre-trained LMs}

Since their recent dominance in standard NLP benchmarks \cite{peters-etal-2018-deep,BERT,RoBERTa}, pre-trained language models have been extensively studied. This has mainly been done through \textit{probing} tasks, which are aimed at understanding the knowledge that is implicitly captured by their parameters. 
After the initial focus on understanding pre-trained LSTM-based LMs \cite{peters-etal:2018-dissecting}, attention has now shifted toward transformer-based models.  The main aspects that have been studied in recent years are syntax \cite{goldberg2019assessing,saphra-lopez:2019,hewitt-manning:2019,van-schijndel-etal-2019-quantity,jawahar-etal-2019-bert,Tenney-et-al-2019} and semantics \cite{Ettinger2019WhatBI,tenney-etal-2019-bert}. For a more complete overview on analyses of the different properties of  transformer-based LMs, we refer to \newcite{rogers2021primer}.

Despite the rise in probing analyses for LMs and the importance of analogical reasoning in human cognition, understanding the analogical capabilities of LMs remains understudied. The most similar works have focused on capturing relational knowledge from LMs (in particular the type of information available in knowledge graphs). For instance, \newcite{petroni2019language} analyzed to what extent LMs could fill manually-defined templates such as \textit{``Dante was born in [MASK]''}. Follow-up works extended this initial approach by automatically generating templates and fine-tuning LMs on them \cite{bouraoui2020inducing,jiang-etal-2020-know}, showing an improved performance. In this paper, we focus on the analogical knowledge that is encoded in pre-trained LMs, without the extra step of fine-tuning on additional data.

\subsection{Word Analogy Probing}
\label{relatedword_analogies}
Word analogies have been used as a standard intrinsic evaluation task for measuring the quality of word embeddings. \newcite{mikolov2013linguistic} showed that word embeddings, in particular Word2vec embeddings, were able to solve analogy problems by simple vector operations (e.g. king - man + woman = queen). The motivation for this task dates back to the connectionism theory \cite{feldman1982connectionist} in cognitive science. In particular, neural networks were thought to be able to model emergent concepts \cite{hopfield1982neural, hinton1986learning}
by learning distributed representations across an embedding space \cite{hinton1986distributed}, similar to the properties that word embeddings displayed in the analogy task. More recent works have proposed new mathematical theories and experiments to understand the analogical capabilities of word embeddings, attempting to understand their linear algebraic structure \cite{arora2016latent,gittens2017skip,allen2019analogies} or by explicitly studying their compositional nature \cite{levy2014linguistic,paperno2016whole,ethayarajh2019towards,chiang-etal-2020-understanding}.

However, recent works have questioned the impressive results displayed by word embeddings in this task. In many cases simple baselines excluding the input pair (or \textit{query}) were competitive \cite{linzen2016issues}. Simultaneously, some researchers have found that many relationships may not be retrieved in the embedding space by simple linear transformations \cite{drozd2016word,bouraoui2018relation} and others argued that the standard evaluation procedure has limitations \cite{schluter2018word}. New datasets and measures have also been introduced to address some of these issues \cite{DBLP:conf/naacl/GladkovaDM16,fournier-etal-2020-analogies}. Finally, in the context of bias detection, for which analogies have been used as a proxy \cite{bolukbasi2016man}, it has also been found that word analogies may misguide or hide the real relationships existing in the vector space  \cite{gonen2019lipstick,nissimCL2020}. 

As far as language models are concerned, word analogies have not been explored to the same extent as for word embeddings. Recently, \newcite{DBLP:conf/nips/BrownMRSKDNSSAA20} evaluated the unsupervised capabilities of GPT-3 by evaluating it on the SAT analogies dataset \cite{DBLP:conf/ranlp/TurneyLBS03}, which we also include in our evaluation (see Section \ref{datasets}). However, the evaluation is limited to a single dataset (i.e., SAT) and model (i.e., GPT-3), and the general capabilities of language models were not investigated.

Despite their limitations, analogy tests remain appealing for evaluating the ability of embeddings and language models to identify abstract relationships.
To mitigate the aforementioned methodological issues, in this work we rely on analogy tests from educational resources, where the task is to complete analogical proportions, given only the first word pair. In contrast, word embedding models have mostly been evaluated using a predictive task, in which three of the four words are given.
Moreover, the considered datasets are focused on abstract analogies, whereas the most commonly used datasets only include well-defined semantic relations such as ``capital of''. For completeness, however, we also show results on these standard datasets. We furthermore experiment with several simple baselines to understand possible artifacts present in the different datasets.

\section{Word Analogies}
\label{analogical_prop}

In this section, we describe the word analogy formulation that is used for our experiments (Section \ref{analogy_task}). Subsequently, we provide an overview of the datasets used in our experiments (Section \ref{datasets}).

\subsection{Task Description}
\label{analogy_task}

We frame the analogy task in terms of analogical proportions \cite{prade2017analogical}. Given a query word pair $(h_q, t_q)$ and a list of candidate answer pairs $\{(h_i, t_i)\}_{i=1}^{n}$, the goal is to find the candidate answer pair that has the most similar relation to the query pair.
Table~\ref{tab:sat-sample} shows a sample query and candidate answers drawn from one of the datasets used in our evaluation (see Section \ref{datasets}).

\subsection{Analogy Datasets}
\label{datasets}

We split analogy datasets in two types, based on how the analogy problems were constructed. 

\subsubsection{Psychometric Analogy Tests} 
Word analogy tests are commonly used in assessments of linguistic and cognitive ability. For instance, in the past,  such tests were included in the SAT exams, which are a US college admission test. \citet{DBLP:conf/ranlp/TurneyLBS03} collected a benchmark of 374 word analogy problems, consisting primarily of problems from these SAT tests. Aimed at college applicants, these problems are designed to be challenging for humans. A key challenge for NLP systems is that solving these problems often requires identifying fine-grained semantic differences between word pairs that belong to the same coarse-grained relation. For instance, in the case of Table \ref{tab:sat-sample}, we could say that ``a year consists of weeks'' like ``language consists of words'', but the \textit{week}-\textit{year} pair is nonetheless less similar to \textit{word}-\textit{language} than \textit{note}-\textit{music}.

Another analogy benchmark was constructed by \citet{boteanu2015solving}, who used word analogy problems from an educational resource\footnote{\url{https://www.englishforeveryone.org/Topics/Analogies.html}}. They used in particular UNIT 2 of the analogy problems from the educational site. These problems have the same form as those from the SAT benchmark, but rather than college applicants, they are aimed at children in grades 4 to 12 from the US school system (i.e.\ from age 9 onwards). In this paper, we will also include this UNIT 2 benchmark. Moreover, we have collected another benchmark from the UNIT 4 problems on the same website. These UNIT 4 problems are organised in 5 difficulty levels: high-beginning, low-intermediate, high-intermediate, low-advanced and high-advanced. The low-advanced level is stated to be at the level of the SAT tests, whereas the high-advanced level is stated to be at the level of the GRE test (which is used for admission into graduate schools).

\begin{table}[t]
\centering
\footnotesize
\scalebox{1.07}{
\begin{tabular}{llcc}
\toprule
\multirow{2}{*}{\textbf{Dataset}}&{\textbf{Data size}} & {\textbf{No.}} & {\textbf{No.}} \\
         & \textbf{(val / test)} & {\textbf{candidates}} & \textbf{groups} \\ 
         \midrule
{SAT}    & 37 / 337            & 5           & 2              \\ 
{UNIT 2} & 24 / 228            & 5,4,3       & 9              \\ 
{UNIT 4} & 48 / 432            & 5,4,3       & 5              \\ 
{Google} & 50 / 500            & 4           & 2              \\ 
{BATS}   & 199 / 1799          & 4           & 3              \\ \bottomrule
\end{tabular}
}
\caption{\label{tab:data}
High-level statistics of the analogy datasets after unification: data size, number of candidates and number of group partitions.
}
\end{table}

\subsubsection{Lexical Semantics Benchmarks}
Since the introduction of Word2vec \cite{mikolov2013distributed}, the problem of modelling analogies has been commonly used as an intrinsic benchmark for word embedding models. However, the datasets that have been used in that context are focused on well-defined and relatively coarse-grained relations. 
The Google analogy dataset \cite{mikolov2013linguistic} has been one of the most commonly used benchmarks for intrinsic evaluation of word embeddings. This dataset contains a mix of semantic and morphological relations such as \textit{capital-of} and \textit{singular-plural}, respectively. However, its coverage has been shown to be limiting, and BATS \cite{DBLP:conf/naacl/GladkovaDM16} was developed in an attempt to address its main shortcomings. BATS includes a larger number of concepts and relations, which are split into four categories: lexicographic, encyclopedic, and derivational and inflectional morphology.

As pointed out above, these datasets were tailored to the evaluation of word embeddings in a predictive setting. To provide an evaluation setting which is comparable to the benchmarks obtained from human analogy tests, we constructed word analogy problems from the Google and BATS datasets, by choosing for each correct analogy pair a number of negative examples. The resulting benchmark thus follows the same format as described in Section \ref{analogy_task}. To obtain sufficiently challenging negative examples, for each query pair (e.g. \textit{Paris-France}) we extracted three negative instances: 
    (1) two random words from the head of the input relation type (e.g. \textit{Rome-Oslo}); 
    (2) two random words from the tail of the input relation type (e.g. \textit{Germany-Canada}); 
    (3) a random word pair from a relation type of the same high-level category as the input relation type (e.g. \textit{Argentina-peso}).\footnote{In order to avoid adding various correct answers to the query, we avoided adding negative pairs from all \textit{country-of} type relations, and from similar lexicographic relations in the BATS dataset with more than one relation type, namely antonyms, synonyms, meronyms and hyponyms.}  

\subsubsection{Unification and Statistics}

Table~\ref{tab:data} provides an overview of our datasets. The instances from each dataset are organised into groups. In the case of Google and BATS, these groups refer to the relation types (e.g.\ semantic or morphological in the case of Google). In the case of UNIT 2 and UNIT 4, the groups refer to the difficulty level. For the SAT dataset, we consider two groups, capturing whether the instances come from an actual SAT test or not. 
Finally, we randomly sample 10$\%$ of each group in each dataset to construct a validation set, and regard the remaining data as the test set.

\section{Methodology}

\begin{figure}[t]
    \centering
    \includegraphics[width=\columnwidth]{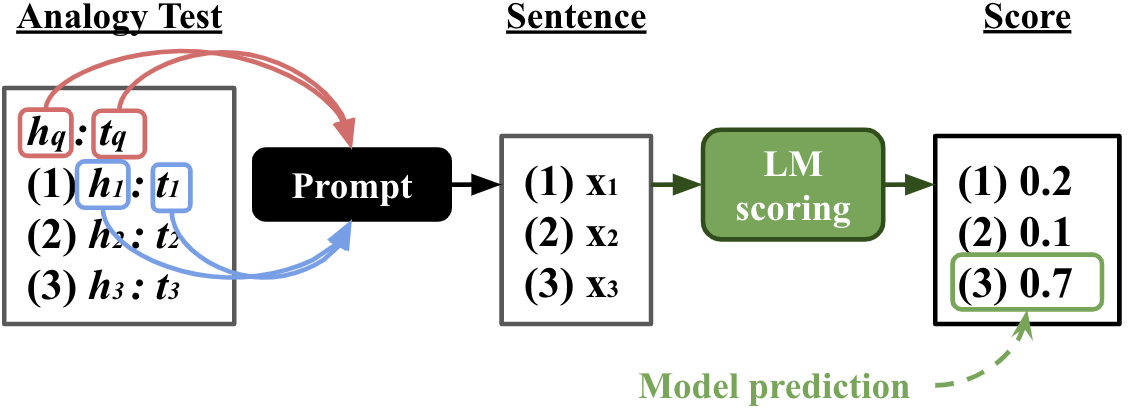}
    \caption{Solving a word analogy problem by selecting one with the highest LM score among the candidates.}
    \label{fig:overview}
\end{figure}

In this section, we explain our 
strategy for using pretrained LMs to solve analogy problems without fine-tuning.
First, in Section \ref{secPrompting} we explain how each relation pair is converted into a natural sentence to be fed into the LM. In Section \ref{secScoring}, we then discuss a number of scoring functions that can be used to select the most plausible answer candidate. Finally, we take advantage of the fact that analogical proportion is invariant to particular permutations, which allows for a natural extension of the proposed scoring functions (Section \ref{permutationinvariance}). 
Figure~\ref{fig:overview} shows a high-level overview of our methodology.

\subsection{Relation Pair Prompting}\label{secPrompting}
We define a prompting function $\mathcal{T}_{t}(w_1, w_2, w_3, w_4)$ that takes four placeholders and a template type $t$, and returns a sentence in which the placeholders were replaced by the words $w_1$, $w_2$, $w_3$, and $w_4$. For instance, given a query ``word:language" and a candidate ``note:music", the prompting function produces
\begin{multline*}
    \mathcal{T}_{\text{{\it to-as}}}(\text{``word"}, \text{``language"}, \text{``note"}, \text{``music"}) =\\
    \text{``word is to language as note is to music"}
\end{multline*}
where we use the template type {\it to-as} here.

Using manually specified template types can result in a sub-optimal textual representation. For this reason, recent studies have proposed auto-prompting strategies, which optimize the template type on a training set \cite{shin-etal-2020-autoprompt}, paraphrasing \cite{jiang-etal-2020-know},
additional prompt generation model \cite{gao2020making},
and corpus-driven template mining \cite{bouraoui2020inducing}. 
However, none of these approaches can be applied to unsupervised settings. 
Thus, we do not explore auto-prompting methods in this work. 
Instead, we will consider a number of different template types in the experiments, and assess the sensitivity of the results to the choice of template type.

\subsection{Scoring Function}\label{secScoring}
\noindent\textbf{Perplexity.}\,
We first define perplexity, which is widely used as a sentence re-ranking metric \cite{chan2016listen, gulcehre2015using}.
Given a sentence ${\boldsymbol x}$, for 
autoregressive LMs such as LSTM based models \cite{zaremba2014recurrent} and GPTs  \cite{radford2018improving,radford2019language,DBLP:conf/nips/BrownMRSKDNSSAA20}, perplexity can be computed as
\begin{equation}\label{eqPPL}
f({\boldsymbol x}) = \exp \left( - \sum_{j=1}^m \log{ P_{\text{auto}}(x_j | {\boldsymbol x}_{j-1} ) } \right)
\end{equation}
where ${\boldsymbol x}$ is tokenized as $[x_1 ... x_m]$ and
$P_{\text{auto}}(x | {\boldsymbol x})$ is the likelihood from an autoregressive LM's next token prediction. For masked LMs such as BERT \cite{BERT} and RoBERTa \cite{RoBERTa}, we instead use pseudo-perplexity, which is defined as in \eqref{eqPPL} but with $P_{\text{mask}}(x_j | {\boldsymbol x}_{\backslash j} )$ instead of $P_{\text{auto}}(x_j | {\boldsymbol x}_{j-1} )$, where ${\boldsymbol x}_{\backslash j} = [x_1 \dots x_{j_1} \text{\textlangle mask\textrangle} x_{j+1} \dots x_m ]$ and $P_{\text{mask}}(x_j | {\boldsymbol x}_{\backslash j})$ is the pseudo-likelihood \cite{wang-cho-2019-bert} that the masked token is $x_j$.

\noindent \textbf{PMI.}\,
Although perplexity is well-suited to capture the fluency of a sentence, it may not be the best choice to test the plausibility of a given analogical proportion candidate. As an alternative, we propose a scoring function that focuses specifically on words from the two given pairs. To this end,
we propose to use an approximation of point-wise mutual information (PMI), based on perplexity.

\begin{figure}[t]
    \centering
    \includegraphics[width=\columnwidth]{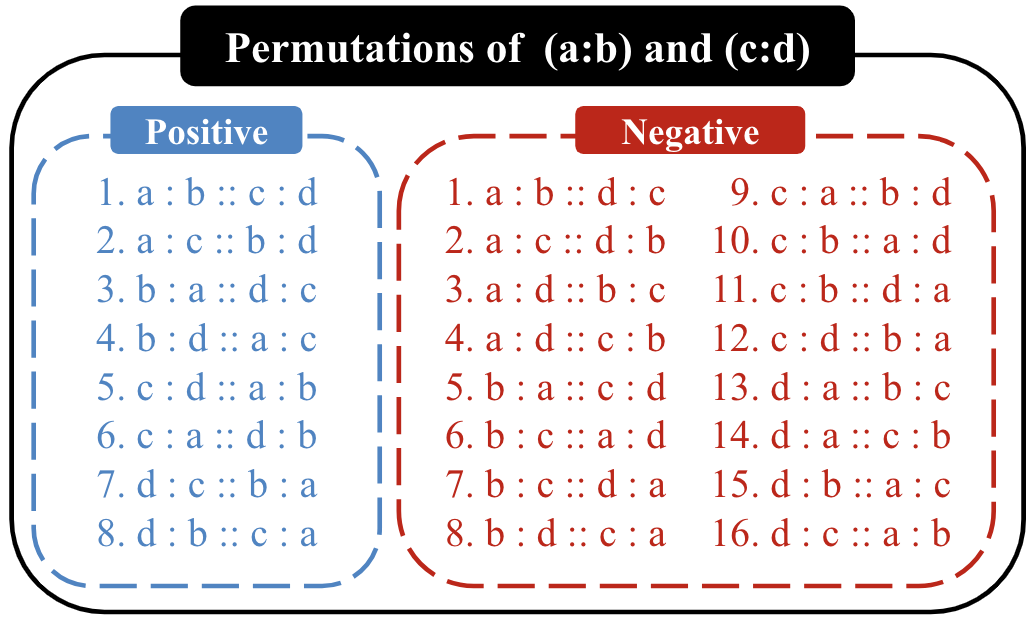}
    \caption{Positive and negative permutations for a relation pair \textit{(a:b)}-\textit{(c:d)}.}
    \label{fig:permutation}
\end{figure}

PMI is defined as the difference between a conditional and marginal log-likelihood.
In our case, we consider the conditional likelihood of $t_i$ given $h_i$ and the query pair (recall from Section \ref{analogy_task} that $h$ and $t$ represent the head and tail of a given word pair, respectively), i.e.\ $P(t_i|h_q, t_q, h_i)$, and the marginal likelihood over $h_i$, i.e. $P(t_i|h_q, t_q)$. Subsequently, the PMI-inspired scoring function is defined as
\begin{align}\label{eqWeightedPMI}
    r(t_i|h_i, h_q, t_q) = & \log P(t_i|h_q, t_q, h_i) \nonumber \\
    &- \alpha \cdot \log P(t_i|h_q, t_q)
\end{align}
where $\alpha$ is a hyperparameter to control the effect of the marginal likelihood. The PMI score corresponds to the specific case where $\alpha=1$. However, \citet{davison2019commonsense} found that using a hyperparameter to balance the impact of the conditional and marginal probabilities can significantly improve the results. 
The probabilities in \eqref{eqWeightedPMI} are estimated by assuming that the answer candidates are the only possible word pairs that need to be considered. By relying on this closed-world assumption, we can estimate marginal probabilities based on perplexity, which we found to give better results than the masking based strategy from \citet{davison2019commonsense}.
In particular, we estimate these probabilities as
\begin{align*}
    P(t_i|h_q, t_q, h_i) &= - \frac{
    f\left(\mathcal{T}_t(h_q, t_q, h_i, t_i) \right)
    }{
    \sum\limits_{k=1}^n f\left(\mathcal{T}_t(h_q, t_q, h_i, t_k) \right)
    } \\
    P(t_i|h_q, t_q) &= - \dfrac{
    \sum\limits_{k=1}^n f\left(\mathcal{T}_t(h_q, t_q, h_k, t_i) \right)
    }{
    \sum\limits_{k=1}^n \sum\limits_{l=1}^n f\left(\mathcal{T}_t(h_q, t_q, h_k, t_l) \right)
    } 
\end{align*}
where $n$ is the number of answer candidates for the given query. Equivalently, since PMI is symmetric, we can consider the difference between the logs of $P(h_i|h_q, t_q, t_i)$ and $P(h_i|h_q, t_q)$. While this leads to the same PMI value in theory, due to the way in which we approximate the probabilities, this symmetric approach will lead to a different score. We thus combine both scores with an aggregation function $\mathcal{A}_{g}$.
This aggregation function takes a list of scores and outputs an aggregated value. As an example, given a list $[1, 2, 3, 4]$, we write $\mathcal{A}_{\text{mean}}([1, 2, 3, 4]) = 2.5$ for the mean and $\mathcal{A}_{\text{val}_1}([1, 2, 3, 4]) = 1$ for the first element.
Given such an aggregation function, we define the following PMI-based score 
\begin{equation}
    s_{\textit{PMI}}(t_i, h_i|h_q, t_q) = \mathcal{A}_g\left( \boldsymbol{r} \right) \label{eq:ppl_pmi}
\end{equation}
where 
we consider basic aggregation operations
over the list $\boldsymbol{r} =  \left[ r(t_i|h_i, h_q, t_q), r(h_i|t_i, h_q, t_q) \right]$,
such as the mean, max, and min value. The choice of using only one of the scores $r(t_i|h_i, h_q, t_q)$, $r(h_i|t_i, h_q, t_q)$ is viewed as a special case, in which the aggregation function $g$ simply returns the first or the second item.

\noindent\textbf{mPPL.}\, We also experiment with a third scoring function, which borrows ideas from both perplexity and PMI. In particular, we propose the marginal likelihood biased perplexity (mPPL) defined as
\begin{align*}
s_{\textit{mPPL}}(t_i, h_i|h_q, t_q) = &\log s_{\textit{PPL}}(t_i, h_i|h_q, t_q)\\
&- \alpha_t \cdot \log P(t_i|h_q, t_q) \\
&- \alpha_h \cdot \log P(h_i|h_q, t_q)
\end{align*}
where $\alpha_t$ and $\alpha_h$ are hyperparameters, and ${s_{\textit{PPL}}}$ is a normalized perplexity defined as
\begin{align*}
    s_{\textit{PPL}}(t_i, h_i|h_q, t_q) &= - \frac{
    f\left(\mathcal{T}_t(h_q, t_q, h_i, t_i) \right)
    }{
    \sum\limits_{k=1}^n f\left(\mathcal{T}_t(h_q, t_q, h_k, t_k) \right)
    }.
\end{align*}
The mPPL score extends perplexity with two bias terms. It is motivated from the insight that treating $\alpha$ as a hyperparameter in \eqref{eqWeightedPMI} can lead to better results than fixing $\alpha=1$. By tuning $\alpha_t$ and $\alpha_h$, we can essentially influence to what extent answer candidates involving semantically similar words to the query pair should be favored. 

\subsection{Permutation Invariance}
\label{permutationinvariance}

The formalization of analogical proportions dates back to Aristotle \cite{barbot2019analogy}. According to the standard axiomatic characterization, whenever we have an analogical proportion $a:b::c:d$ (meaning ``$a$ is to $b$ what $c$ is to $d$''), it also holds that $c:d::a:b$ and $a:c::b:d$ are analogical proportions. It follows from this that for any given analogical proportion $a:b::c:d$ there are eight permutations of the four elements $a,b,c,d$ that form analogical proportions. These eight permutations, along with the 16 ``negative permutations'', are shown in Figure~\ref{fig:permutation}.

To take advantage of the different permutations of analogical proportions, we propose the following Analogical Proportion (AP) score:
\begin{align}
    \label{eq:rel-pmi}
    \textit{AP}(h_q, t_q, h_i, t_i) &= \mathcal{A}_{g_{\text{pos}}}(\boldsymbol{p}) - \beta \cdot \mathcal{A}_{g_{\text{neg}}}(\boldsymbol{n})\\
    \boldsymbol{p} &= \left[ s({a}, {b}| {c}, {d}) \right]_{({a}:{b}, {c}:{d}) \in \boldsymbol{P}} \notag \\
    \boldsymbol{n} &= \left[ s({a}, {b}| {c}, {d}) \right]_{({a}:{b}, {c}:{d}) \in \boldsymbol{N}}\notag
\end{align}
where $\boldsymbol{P}$ and $\boldsymbol{N}$ correspond to the list of positive and negative permutations of the candidate analogical proportion $h_q:t_q::h_i: t_i$ in the order shown in Figure~\ref{fig:permutation}, 
$\beta$ is a hyperparameter to control the impact of the negative permutations,
and $s(a, b | c, d)$ is a scoring function as described in Section~\ref{secScoring}.
Here $\mathcal{A}_{g_{\text{pos}}}$ and $\mathcal{A}_{g_{\text{neg}}}$ refer to the aggregation functions that are used to combine the scores for the positive and negative permutations respectively, where these aggregation functions are defined as in Section~\ref{secScoring}.
To solve an analogy problem, we simply choose the answer candidate that results in the highest value of $\textit{AP}(t_i, h_i, h_q, t_q)$.

\section{Evaluation}
In this section, we evaluate language models on the five analogy datasets presented in Section \ref{analogical_prop}.

\begin{table*}[t]
\centering
\setlength{\tabcolsep}{6.0pt}
\begin{tabular}{cccc@{\hspace{30pt}}C{3em}@{}C{3em}@{}C{3em}@{}C{3em}@{}C{3em}@{\hspace{30pt}}c}
\toprule
\multicolumn{2}{c}{\textbf{Model}}  & \textbf{Score}    & \textbf{Tuned}& \textbf{SAT}     & \textbf{U2}   & \textbf{U4}   & \textbf{Google}   & \textbf{BATS}  & \textbf{Avg} \\\hline
\multirow{15}{*}{\rotatebox{90}{LM}} 
& \multirow{5}{*}{BERT} & \multirow{2}{*}{$s_{\textit{PPL}}$} & & 32.9 & 32.9 & 34.0 & 80.8 & 61.5 & 48.4 \\
& & & \checkmark & 39.8 & 41.7 & 41.0 & 86.8 & 67.9 & 55.4 \\\cline{3-10} 
& & \multirow{2}{*}{$s_{\textit{PMI}}$} & & 27.0 & 32.0 & 31.2 & 74.0 & 59.1 & 44.7 \\
& & & \checkmark & 40.4 & 42.5 & 27.8 & 87.0 & 68.1 & 53.2 \\\cline{3-10}
& & $s_{\textit{mPPL}}$ & \checkmark & 41.8 & 44.7 & 41.2 & 88.8 & 67.9 & 56.9 \\\cline{2-10} 
& \multirow{5}{*}{GPT-2} & \multirow{2}{*}{$s_{\textit{PPL}}$} & & 35.9 & 41.2 & 44.9 & 80.4 & 63.5 & 53.2 \\
& & & \checkmark & 50.4 & 48.7 & 51.2 & 93.2 & 75.9 & 63.9 \\\cline{3-10} 
& & \multirow{2}{*}{$s_{\textit{PMI}}$} & & 34.4 & 44.7 & 43.3 & 62.8 & 62.8 & 49.6 \\
& & & \checkmark & 51.0 & 37.7 & 50.5 & 91.0 & 79.8 & 62.0 \\\cline{3-10}
& & {$s_{\textit{mPPL}}$} & \checkmark & \textbf{56.7} & 50.9 & 49.5 & 95.2 & \textbf{81.2} & 66.7 \\\cline{2-10}
& \multirow{5}{*}{RoBERTa} & \multirow{2}{*}{$s_{\textit{PPL}}$} & & 42.4 & 49.1 & 49.1 & 90.8 & 69.7 & 60.2 \\
& & & \checkmark & 53.7 & 57.0 & 55.8 & 93.6 & 80.5 & 68.1 \\\cline{3-10} 
& & \multirow{2}{*}{$s_{\textit{PMI}}$} & & 35.9 & 42.5 & 44.0 & 60.8 & 60.8 & 48.8 \\
& & & \checkmark & 51.3 & 49.1 & 38.7 & 92.4 & 77.2 & 61.7 \\ \cline{3-10}
& & $s_{\textit{mPPL}}$ & \checkmark & 53.4 & \textbf{58.3} & \textbf{57.4} & 93.6 & 78.4 & \textbf{68.2} \\ \cline{1-10} 
\multirow{3}{*}{\rotatebox{90}{WE}} & FastText  & -                     &               & 47.8              & 43.0          & 40.7          & \textbf{96.6}     & 72.0 & 60.0\\ 
                    & GloVe                     & -                     &               & 47.8              & 46.5          & 39.8          & 96.0              & 68.7 & 59.8 \\ 
                    & Word2vec                  & -                     &               & 41.8              & 40.4          & 39.6          & 93.2              & 63.8 & 55.8  \\ \hline
\multirow{2}{*}{\rotatebox{90}{Base\:}}   &   PMI & -                     &               & 23.3              & 32.9          & 39.1          & 57.4              & 42.7 & 39.1 \\
                    & Random                    & -                     &               & 20.0              & 23.6          & 24.2          & 25.0              & 25.0 & 23.6 \\ 
\bottomrule
\end{tabular}
\caption{\label{tab:main-result}
Accuracy results on each analogy dataset, categorized into language models (LM), word embeddings (WE), and baselines (Base). All LMs use the analogical proportion (AP) function described in Section \ref{permutationinvariance}. The default configuration for AP includes $\alpha=\alpha_h=\alpha_t=\beta=0$, $g_{\text{pos}}=g=\text{val}_{1}$, and $t=\text{{\it to-as}}$. Note that $s_{\textit{PPL}}=s_{\textit{mPPL}}$ with the default configuration.
Average accuracy (Avg) across datasets is included in the last column.
}
\end{table*}

\begin{table}[t]
\centering
\scalebox{0.85}{
\begin{tabular}{ccccc}
\toprule
                            & \textbf{Model}            & \textbf{Score}                        & \textbf{Tuned}    & \textbf{Accuracy} \\\hline
\multirow{17}{*}{LM}        & \multirow{5}{*}{BERT}     & \multirow{2}{*}{$s_{\textit{PPL}}$}   &                   & 32.6~ \\ 
                            &                           &                                       & \checkmark        & 40.4* \\\cline{3-5} 
                            &                           & \multirow{2}{*}{$s_{\textit{PMI}}$}   &                   & 26.8~ \\
                            &                           &                                       & \checkmark        & 41.2* \\ \cline{3-5} 
                            &                           & $s_{\textit{mPPL}}$                   & \checkmark        & 42.8* \\ \cline{2-5} 
                            & \multirow{5}{*}{GPT-2}    & \multirow{2}{*}{$s_{\textit{PPL}}$}   &                   & 41.4~ \\ 
                            &                           &                                       & \checkmark        & 56.2*\\\cline{3-5} 
                            &                           & \multirow{2}{*}{$s_{\textit{PMI}}$}   &                   & 34.7~ \\
                            &                           &                                       & \checkmark        & 56.8* \\ \cline{3-5} 
                            &                           & $s_{\textit{mPPL}}$                    & \checkmark        & 57.8* \\ \cline{2-5} 
                            & \multirow{5}{*}{RoBERTa}  & \multirow{2}{*}{$s_{\textit{PPL}}$}   &                   & 49.6~ \\ 
                            &                           &                                       & \checkmark        & 55.8* \\ \cline{3-5} 
                            &                           & \multirow{2}{*}{$s_{\textit{PMI}}$}   &                   & 42.5~ \\
                            &                           &                                       & \checkmark        & 54.0* \\ \cline{3-5} 
                            &                           & $s_{\textit{mPPL}}$                   & \checkmark        & 55.8* \\ \cline{2-5} 
                            & \multirow{2}{*}{GPT-3}    & \textit{Zero-shot}&                   & \textit{53.7~}     \\
                            &                           & \textit{Few-shot} & \checkmark        & \textit{65.2}*     \\ \hline
                        -   & LRA                       & -                 & & \textit{56.4~}     \\ \hline
\multirow{3}{*}{WE}         & FastText                  & -                 & & 49.7~     \\
                            & GloVe                     & -                 & & 48.9~     \\
                            & Word2vec                  & -                 & & 42.8~     \\ \hline

\multirow{2}{*}{Base}      & PMI                        & -                 & & 23.3~     \\
                           & Random                     & -                 & & 20.0~     \\ \bottomrule
\end{tabular}
}
\caption{
\label{tab:sat-full}
Accuracy results for the full SAT dataset. Results marked with * are not directly comparable as they were tuned on full data (for our models) or use training data (for GPT-3 few-shot). These results are included to provide an upper bound only. Results in italics were taken from the original papers.
}

\end{table}

\subsection{Experimental Setting}

We consider three transformer-based LMs of a different nature: two masked LMs, namely BERT \cite{BERT} and RoBERTa \cite{RoBERTa}, and GPT-2, as a prominent example of an auto-regressive language model.
Each pretrained model was fetched from the Huggingface transformers library \citep{Wolf2019HuggingFacesTS}, from which we use \texttt{bert-large-cased}, \texttt{roberta-large}, and \texttt{gpt2-xl} respectively. For parameter selection, we run grid search on $\beta$, $\alpha$, $\alpha_h$, $\alpha_t$, $t$, $g$, $g_{\text{{\it pos}}}$, and $g_{\text{{\it neg}}}$ for each model and select the configuration which achieves the best accuracy on each validation set. We experiment with the three scoring functions presented in Section \ref{secScoring}, i.e., $s_{\textit{PPL}}$ (perplexity), $s_{\textit{PMI}}$ and $s_{\textit{mPPL}}$.
Possible values for each hyperparameter (including the selection of six prompts and an ablation test on the scoring function) and the best configurations that were found by grid search are provided in the appendix.

As baseline methods, we also consider three pre-trained word embedding models, which have been shown to provide competitive results in analogy tasks, as explained in Section \ref{relatedword_analogies}: Word2vec \cite{mikolov2013distributed}, GloVe \cite{pennington2014glove}, and FastText \cite{bojanowski2017enriching}. For the word embedding models, we simply represent word pairs by taking the difference between their embeddings\footnote{Vector differences have been found to be the most robust encoding method in the context of word analogies \cite{hakami2017compositional}.}. We then choose the answer candidate with the highest cosine similarity to the query in terms of this vector difference.  To put the results into context, we also include two simple statistical baselines. First, we report the expected random performance. Second, we use a method based on each word pair's PMI in a given corpus. We then select the answer candidate with the highest PMI as the prediction. Note that the query word pair is completely ignored in this case. This PMI score is the well-known word-pair association metric introduced by \citet{church1990word} for lexicographic purposes (specifically, collocation extraction), which compares the probability of observing two words together with the probabilities of observing them independently (chance). The PMI scores in our experiments were computed using the English Wikipedia with a fixed window size 10.

\subsection{Results}
\label{results}

Table~\ref{tab:main-result} shows our main results. 
As far as the comparison among LMs is concerned, RoBERTa and GPT-2 consistently outperform BERT.
Among the AP variants, $s_{\textit{mPPL}}$ achieves substantially better results than $s_{\textit{PMI}}$ or $s_{\textit{PPL}}$ in most cases. 
We also observe that word embeddings perform surprisingly well, with FastText and GloVe outperforming BERT on most datasets, as well as GPT-2 and RoBERTa with default hyperparameters. 
FastText achieves the best overall accuracy on the Google dataset, confirming that this dataset is particularly well-suited to word embeddings (see Section \ref{relatedword_analogies}).

In order to compare with published results from prior work, we carried out an additional experiment on the full SAT dataset (i.e., without splitting it into validation and test). Table~\ref{tab:sat-full} shows the results. GPT-3 \cite{DBLP:conf/nips/BrownMRSKDNSSAA20} and LRA \cite{Turney:2005:MSS:1642293.1642475} are added for comparison. Given the variability of the results depending on the tuning procedure, we have also reported results of configurations that were tuned on the entire set, to provide an upper bound on what is possible within the proposed unsupervised setting. This result shows that even with optimal hyperparameter values, LMs barely outperform the performance of the simpler LRA model. GPT-3 similarly fails to outperform LRA in the zero-shot setting.

\section{Analysis}
We now take a closer look into our results to investigate parameter sensitivity, the correlation between model performance and human difficulty levels, and possible dataset artifacts.
The following analysis focuses on $s_{\textit{mPPL}}$ as it achieved the best results among the LM based scoring functions.

\paragraph{Parameter Sensitivity}
\begin{figure}[t]
    \centering
    \includegraphics[width=\columnwidth]{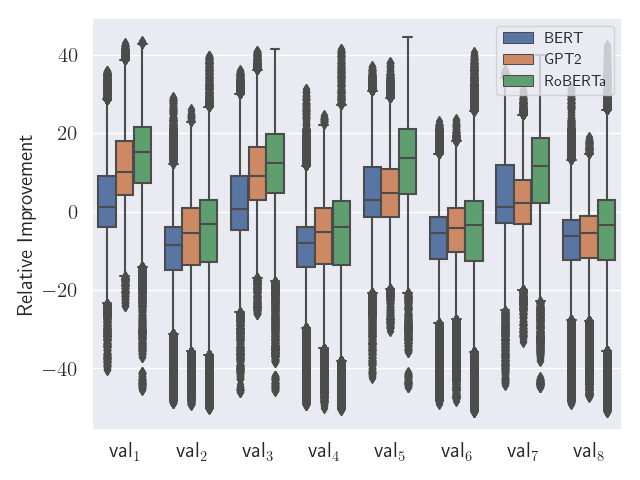}
    \caption{Box plot of the relative improvement on test accuracy in each dataset over all configurations of $s_{\textit{mPPL}}$ grouped by $g_{\text{pos}}$. Here val$_{k}$ corresponds to $k$th positive permutation shown in Figure~\ref{fig:permutation}.}
    \label{fig:box}
\end{figure}
We found that optimal values of the parameters $\alpha$ and $\beta$ are highly dependent on the dataset, while other parameters such as the template type $t$ vary across LMs. On the other hand, as shown in Figure~\ref{fig:box}, the optimal permutations of the templates are relatively consistent, with the original ordering $a:b::c:d$ typically achieving the best results. The results degrade most for permutations that mix the two word pairs (e.g.\ $a:c::b:d$). In the appendix we include an ablation study for the sensitivity and relevance of other parameters and design choices.

\paragraph{Difficulty Levels}
\label{analysis:difficulty}
\begin{figure}[t]
    \centering
    \includegraphics[width=\columnwidth]{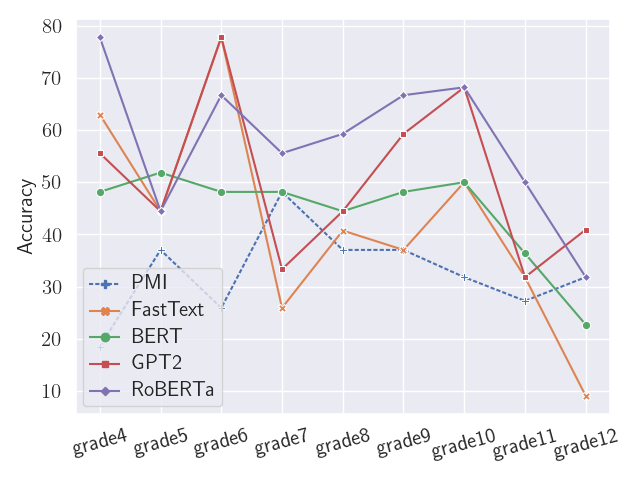}
    \centering
    \includegraphics[width=\columnwidth]{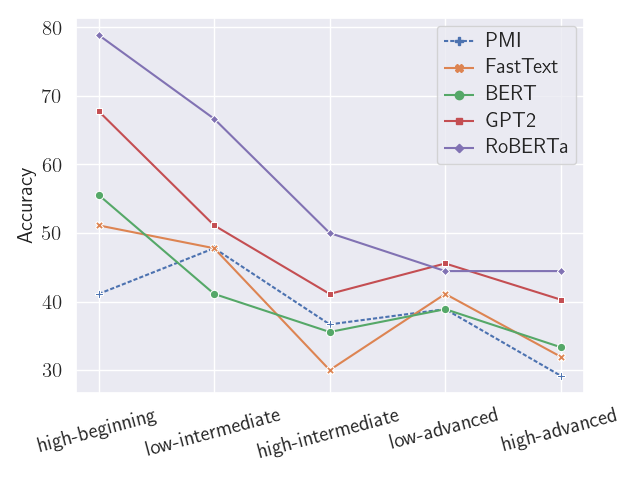}
    \caption{Test accuracy in U2 and U4 per difficulty level. LMs use $s_{\textit{mPPL}}$ with the best configuration tuned in the corresponding validation sets.}
    \label{fig:line-u2}
\end{figure}
To increase our understanding of what makes an analogy problem difficult for LMs, we compare the results for each difficulty level.\footnote{For SAT, Google and BATS, there are no difficulty levels available, but we show the results split by high-level categories in the appendix. We also note that the number of candidates in U2 and U4 vary from three to five, so results per difficulty level are not fully comparable. However, they do reflect the actual difficulty of the educational tests.} Recall from Section \ref{datasets} that the U2 and U4 datasets come from educational resources and are split by difficulty level. Figure \ref{fig:line-u2} shows the results of all LMs (tuned setting), FastText and the PMI baseline according to these difficulty levels. Broadly speaking, we can see that instances that are harder for humans are also harder for the considered models. The analogies in the most difficult levels are generally more abstract (e.g.\ $\textit{witness}:\textit{testimony}::\textit{generator}:\textit{electricity}$), or contain obscure or infrequent words (e.g.\ $\textit{grouch}:\textit{cantakerous}::\textit{palace}:\textit{ornate}$).\footnote{In the appendix we include more examples with errors made by RoBERTa in \textit{easy} instances.}

\paragraph{Hypothesis Only}

Recently, several researchers have found that standard NLP benchmarks, such as SNLI \cite{bowman-etal-2015-large} for language inference, contain several annotation artifacts that makes the task simpler for automatic models \cite{poliak-etal-2018-hypothesis,gururangan-etal-2018-annotation}. One of their most relevant findings is that models which do not even consider the premise can reach high accuracy. More generally, these issues have been found to be problematic in NLP models \cite{linzen-2020-accelerate} and neural networks more generally \cite{geirhos2020shortcut}. According to the results shown in Table~\ref{tab:main-result}, we already found that the PMI baseline achieved a non-trivial performance, even outperforming BERT in a few settings and datasets. This suggests that several implausible negative examples are included in the analogy datasets. As a further exploration of such artifacts, here we analyse the analogue of a \textit{hypothesis-only} baseline. In particular, for this analysis, we masked the head or tail of the candidate answer in all evaluation instances. Then, we test the masked language models with the same AP configuration and tuning on these artificially-modified datasets.
As can be seen in Table \ref{table:hypothesisonly}, a non-trivial performance is achieved for all datasets, which suggests that the words from the answer pair tend to be more similar to the words from the query than the words from negative examples.

\begin{table}[t]
\centering
\footnotesize
\scalebox{1.05}{
\begin{tabular}{ll@{\hspace{7pt}}c@{\hspace{7pt}}c@{\hspace{7pt}}c@{\hspace{7pt}}c@{\hspace{7pt}}c}
\toprule
                                               & \textbf{Mask}& \textbf{SAT} & \textbf{U2} & \textbf{U4} & \textbf{Google} & \textbf{BATS} \\ \midrule
\multirow{3}{*}{\rotatebox{90}{{\scriptsize BERT}}} & full         & 41.8         & 44.7        & 41.2        & 88.8            & 67.9 \\ 
                                               & head         & 31.8         & 28.1        & 34.3        & 72.0            & 62.4 \\ 
                                               & tail         & 33.5         & 31.6        & 38.2        & 64.2            & 63.1  \\  \midrule
 \multirow{3}{*}{\rotatebox{90}{{\scriptsize	 RoBERTa}}}& full         & 53.4         & 58.3        & 57.4        & 93.6            & 78.4 \\
                                               & head         & 38.6         & 37.7        & 41.0        & 60.6            & 54.5  \\ 
                                               & tail         & 35.6         & 37.3        & 40.5        & 55.8            & 64.2 \\ \bottomrule
\end{tabular}
}
\caption{Accuracy results by masking head or tail of the candidate answers. Results in the top row correspond to the full model without masking. \label{table:hypothesisonly}}
\end{table}

\section{Conclusion}

In this paper, we have presented an extensive analysis of the ability of language models to identify analogies. To this end, we first compiled datasets with psychometric analogy problems from educational resources, covering a wide range of difficulty levels and topics. We also recast two standard benchmarks, the Google and BATS analogy datasets, into the same style of problems. Then, we proposed standard techniques to apply language models to the unsupervised task of solving these analogy problems. Our empirical results shed light on the strengths and limitations of various models. To directly answer the question posed in the title, our conclusion is that language models can identify analogies to a certain extent, but not all language models are able to achieve a meaningful improvement over word embeddings (whose limitations in analogy tasks are well documented). On the other hand, when carefully tuned, some language models are able to achieve state-of-the-art results. We emphasize that results are highly sensitive to the chosen hyperparameters (which define the scoring function and the prompt among others). Further research could focus on the selection of these optimal hyperparameters, including automatizing the search or generation of prompts, along the lines of \newcite{bouraoui2020inducing} and \newcite{shin-etal-2020-autoprompt}, respectively.  
Finally, clearly LMs might still be able to learn to solve analogy tasks when given appropriate training data, which is an aspect that we leave for future work.





\bibliographystyle{acl_natbib}
\bibliography{anthology,acl2021}

\appendix

\section{Experimental Details}
In our grid search to find the optimal configuration for each dataset and language model, each parameter was selected within the values shown in Table~\ref{tab:parameters}.
As the coefficient of marginal likelihood $\alpha, \alpha_h, \alpha_t$, we considered negative values as well as we hypothesized that the marginal likelihood could be beneficial for LMs as a way to leverage lexical knowledge of the head and tail words.

Additionally, Table~\ref{tab:template} shows the set of custom templates (or prompts) used in our experiments.  
Finally, Tables~\ref{tab:tuned-parameter-pmi}, \ref{tab:tuned-parameter-mppl},
and \ref{tab:tuned-parameter-hy} include the best configuration based on each validation set in  for $s_{\textit{PMI}}$, $s_{\textit{mPPL}}$ and the hypothesis-only baseline, respectively.

\begin{table}[h]
\centering
\centering
\scalebox{0.85}{
\begin{tabular}{ll}
\toprule
\textbf{Parameter}  & \textbf{Value}                   \\ 
\midrule
$\alpha$            & -0.4, -0.2, 0, 0.2, 0.4    \\ 
$\alpha_h$          & -0.4, -0.2, 0, 0.2, 0.4    \\ 
$\alpha_t$          & -0.4, -0.2, 0, 0.2, 0.4    \\ 
$\beta$             & 0, 0.2, 0.4, 0.6, 0.8, 1.0 \\ 
$g$                 & max,mean,min,val$_1$,val$_{2}$    \\ 
$g_{\text{pos}}$    & max,mean,min,val$_{1}$,...,val$_{8}$ \\ 
$g_{\text{neg}}$    & max,mean,min,val$_{1}$,...,val$_{16}$    \\
\bottomrule
\end{tabular}
}
\caption{\label{tab:parameters}
Hyperparameters with each search space.}
\end{table}

\begin{table}[h]
\centering
\scalebox{0.85}{
\begin{tabular}{ll}
\toprule
{\bf Type} & {\bf Template} \\
\midrule
{\it to-as} & [$w_1$] is to [$w_2$] as [$w_3$] is to [$w_4$] \\
{\it to-what} & [$w_1$] is to [$w_2$] What [$w_3$] is to [$w_4$] \\
\multirow{3}{*}{{\it rel-same}} & The relation between [$w_1$] and [$w_2$] \\
               & is the same as the relation between \\
               & [$w_3$] and [$w_4$]. \\
{\it what-to} & what [$w_1$] is to [$w_2$], [$w_3$] is to [$w_4$] \\
\multirow{2}{*}{{\it she-as}} & She explained to him that [$w_1$] is\\
                        & to [$w_2$] as [$w_3$] is to [$w_4$] \\
\multirow{3}{*}{{\it as-what}} & As I explained earlier, what [$w_1$] is \\
& to [$w_2$] is essentially the same as \\
& what [$w_3$] is to [$w_4$]. \\ \bottomrule
\end{tabular}
}
\caption{\label{tab:template} Custom templates used in our experiments. Each has four placeholders $[w_{1},...,w_{4}]$ and they are fulfilled by words from a relation pair.}
\end{table}

\begin{table}[h]
\centering
\resizebox{\columnwidth}{!}{
\begin{tabular}{ll|rrrrrr}
\toprule
& \textbf{Data} & {$g$}         & {$\alpha$}    & {$g_{\text{pos}}$}    & {$g_{\text{neg}}$}    & {$\beta$} & {$t$}   \\ \midrule
\multirow{5}{*}{\rotatebox{90}{{BERT}}}
& SAT           & val$_2$       & -0.4          & val$_5$               & val$_{12}$            & 0.4       & {\it what-to}  \\
& U2            & val$_2$       & -0.4          & mean                  & mean                  & 0.6       & {\it what-to}  \\
& U4            & val$_1$       & 0.4           & max                   & val$_{7}$             & 1.0       & {\it rel-same} \\
& Google        & val$_1$       & -0.4          & val$_1$               & val$_{11}$            & 0.4       & {\it she-as}  \\
& BATS          & val$_1$       & -0.4          & val$_{11}$            & val$_{1}$             & 0.4       & {\it she-as}  \\ \midrule
\multirow{5}{*}{\rotatebox{90}{{GPT-2}}}
& SAT           & val$_2$       & -0.4          & val$_3$               & val$_{1}$             & 0.6       & {\it rel-same}  \\
& U2            & val$_2$       & 0.0           & val$_{4}$             & val$_{4}$             & 0.6       & {\it rel-same} \\
& U4            & val$_2$       & -0.4          & mean                  & mean                  & 0.6       & {\it rel-same}    \\
& Google        & val$_1$       & 0.0           & mean                  & val$_{11}$            & 0.4       & {\it as-what}  \\
& BATS          & val$_1$       & -0.4          & val$_1$               & val$_{6}$             & 0.4       & {\it rel-same}  \\ \midrule
\multirow{5}{*}{\rotatebox{90}{{RoBERTa}}}
& SAT           & min           & -0.4          & min                   & val$_{7}$             & 0.2       & {\it as-what}  \\
& U2            & min           & 0.4           & mean                  & val$_{4}$             & 0.6       & {\it what-to} \\
& U4            & val$_2$       & 0.0           & mean                  & val$_{4}$             & 0.8       & {\it to-as}    \\
& Google        & val$_1$       & -0.4          & val$_{1}$             & val$_{6}$             & 0.4       & {\it what-to}  \\
& BATS          & max           & -0.4          & mean                  & val$_{11}$            & 0.6       & {\it what-to}  \\ \hline
\end{tabular}
}
\caption{\label{tab:tuned-parameter-pmi}
The best configuration of $s_{\textit{PMI}}$ score.}
\end{table}

\begin{table}[h]
\centering
\resizebox{\columnwidth}{!}{
\begin{tabular}{ll|rrrrrr}
\toprule
& \textbf{Data} & {$\alpha_h$}  & {$\alpha_t$}  & {$g_{\text{pos}}$}    & {$g_{\text{neg}}$}    & {$\beta$} & {$t$} \\ \midrule
\multirow{5}{*}{\rotatebox{90}{{BERT}}}
& SAT           & -0.2          & -0.4          & val$_5$               & val$_5$               & 0.2       & {\it what-to}\\
& U2            & 0.0           & -0.2          & mean                  & mean                  & 0.8       & {\it she-as} \\
& U4            & -0.2          & 0.4           & val$_7$               & min                   & 0.4       & {\it to-as}    \\
& Google        & 0.4           & -0.2          & val$_5$               & val$_{12}$            & 0.6       & {\it she-as}  \\
& BATS          & 0.0           & 0.0           & val$_8$               & min                   & 0.4       & {\it what-to}  \\ \midrule
\multirow{5}{*}{\rotatebox{90}{{GPT-2}}}
& SAT           & -0.4          & 0.2           & val$_3$               & val$_1$               & 0.8       & {\it rel-same}  \\
& U2            & -0.2          & 0.2           & mean                  & mean                  & 0.8       & {\it as-what} \\
& U4            & -0.2          & 0.2           & mean                  & mean                  & 0.8       & {\it rel-same}  \\
& Google        & -0.2          & -0.4          & mean                  & mean                  & 0.8       & {\it rel-same}  \\
& BATS          & 0.4           & -0.4          & val$_1$               & val$_{5}$             & 0.8       & {\it rel-same}  \\\midrule
\multirow{5}{*}{\rotatebox{90}{{RoBERTa}}}
& SAT           & 0.2           & 0.2           & val$_5$               & val$_{11}$            & 0.2       & {\it as-what}  \\
& U2            & 0.4           & 0.4           & val$_1$               & val$_4$               & 0.4       & {\it what-to} \\
& U4            & 0.2           & 0.2           & val$_1$               & val$_1$               & 0.4       & {\it as-what}  \\
& Google        & 0.2           & 0.2           & val$_1$               & val$_6$               & 0.2       & {\it what-to}  \\
& BATS          & 0.2           & -0.2          & val$_5$               & val$_{11}$            & 0.4       & {\it what-to}  \\\bottomrule
\end{tabular}
}
\caption{\label{tab:tuned-parameter-mppl}
The best configuration of $s_{\textit{mPPL}}$ score.}
\end{table}

\begin{table}[h]
\centering
\scalebox{0.85}{
\begin{tabular}{lllrr}
\toprule
                          & \textbf{Mask}                  & Data   & $g_{\text{{\it pos}}}$ & $t$ \\ \midrule
\multirow{10}{*}{\rotatebox{90}{{ BERT}}}   & \multirow{5}{*}{head} & SAT    & val$_5$                & {\it to-what} \\
                          &                       & U2     & val$_5$                & {\it to-as}   \\
                          &                       & U4     & mean                   & {\it to-as}   \\
                          &                       & Google & val$_5$                & {\it she-as}  \\
                          &                       & BATS   & val$_5$                & {\it to-as}   \\ \cline{2-5}
                          & \multirow{5}{*}{tail} & SAT    & val$_3$                & {\it what-to} \\
                          &                       & U2     & val$_7$                & {\it to-what} \\
                          &                       & U4     & val$_4$                & {\it rel-same}\\
                          &                       & Google & val$_7$                & {\it as-what} \\
                          &                       & BATS   & val$_7$                & {\it to-as}   \\ \midrule
\multirow{10}{*}{\rotatebox{90}{{ RoBERTa}}}& \multirow{5}{*}{head} & SAT    & val$_5$                & {\it as-what} \\
                          &                       & U2     & val$_5$                & {\it rel-same}\\
                          &                       & U4     & val$_7$                & {\it she-as}  \\
                          &                       & Google & val$_5$                & {\it what-to} \\
                          &                       & BATS   & val$_5$                & {\it she-as}  \\\cline{2-5}
                          & \multirow{5}{*}{tail} & SAT    & mean                   & {\it what-to} \\
                          &                       & U2     & val$_7$                & {\it rel-same}\\
                          &                       & U4     & mean                   & {\it what-to} \\
                          &                       & Google & val$_7$                & {\it as-what} \\
                          &                       & BATS   & val$_7$                & {\it what-to} \\\bottomrule
\end{tabular}
}
\caption{
\label{tab:tuned-parameter-hy}
The best configurations for hypothesis-only scores.}
\end{table}

\section{Additional Ablation Results}
We show a few more complementary results to our main experiments.

\subsection{Alternative Scoring Functions}
As alternative scoring functions for LM, we have tried two other scores: PMI score based on masked token prediction \cite{davison2019commonsense} (Mask PMI) and cosine similarity between the embedding difference of a relation pair similar to what used in word-embedding models. For embedding method, we give a prompted sentence to LM to get the last layer's hidden state for each word in the given pair and we take the difference between them, which we regard as the embedding vector for the pair. Finally we pick up the most similar candidate in terms of the cosine similarity with the query embedding. Table~\ref{tab:alternate-scores} shows the test accuracy on each dataset. As one can see, AP scores outperform other methods with a great margin.

\begin{table}[h]
\centering
\scalebox{0.85}{
\begin{tabular}{ll@{\hspace{7pt}}c@{\hspace{7pt}}c@{\hspace{7pt}}c@{\hspace{7pt}}c@{\hspace{7pt}}c}
\toprule
& \textbf{Score}& \textbf{SAT} & \textbf{U2} & \textbf{U4} & \textbf{Google} & \textbf{BATS} \\ \midrule
\multirow{4}{*}{\rotatebox{90}{{\small BERT}}} 
& embedding     & 24.0         & 22.4        & 26.6        & 28.2            & 28.3 \\ 
& Mask PMI      & 25.2         & 23.3        & 31.5        & 61.2            & 46.2 \\ 
& $s_{\textit{PMI}}$        & 40.4          & 42.5      & 27.8          & 87.0          & 68.1 \\ 
& $s_{\textit{mPPL}}$       & 41.8          & 44.7      & 41.2          & 88.8          & 67.9  \\  \midrule
\multirow{4}{*}{\rotatebox{90}{{\small RoBERTa}}}
& embedding     & 40.4         & 42.5        & 27.8        & 87.0            & 68.1 \\ 
& Mask PMI      & 43.0         & 36.8        & 39.4        & 69.2            & 58.3 \\ 
& $s_{\textit{PMI}}$        & 51.3          & 49.1      & 38.7          & 92.4          &  77.2 \\ 
& $s_{\textit{mPPL}}$       & 53.4          & 58.3      & 57.4          & 93.6          & 78.4  \\  \bottomrule
\end{tabular}
}
\caption{
\label{tab:alternate-scores}
Test accuracy tuned on each validation set.}
\end{table}

\subsection{Parameter Sensitivity: template type $t$}
\begin{figure}[h]
    \centering
    \includegraphics[width=\columnwidth]{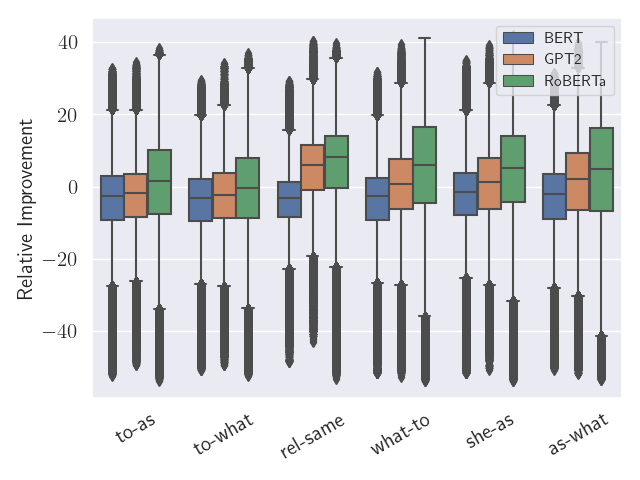}
    \caption{
    Box plot of the relative improvement on test accuracy in each dataset over all configurations of $s_{\textit{mPPL}}$ grouped by template type.}
    \label{fig:box-temp}
\end{figure}

Figure~\ref{fig:box-temp} shows the box plot of relative improvement across all datasets grouped by $t$ and the results indicate that there is a mild trend that certain templates tend to perform well, but not significant universal selectivity can be found across datasets. 

\subsection{Parameter Sensitivity: aggregation method $g_{\text{neg}}$}
\begin{figure}[h]
    \centering
    \includegraphics[width=\columnwidth]{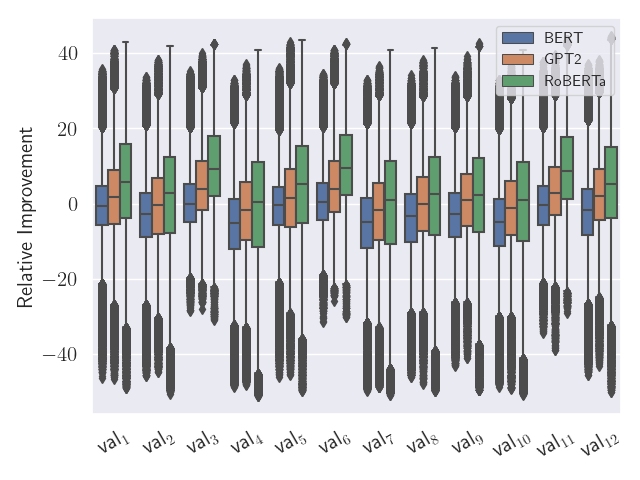}
    \caption{
    Box plot of the relative improvement on test accuracy in each dataset over all configurations of $s_{\textit{mPPL}}$ grouped by $g_{\text{neg}}$. Here val$_{k}$ corresponds to $k$th positive permutation shown in Figure~\ref{fig:permutation}.}
    \label{fig:negative_perm}
\end{figure}

Figure~\ref{fig:negative_perm} shows the box plot of relative improvement  across all datasets grouped by $g_{\text{neg}}$.
Unlike $g_{\text{pos}}$ we show in Figure~\ref{fig:box}, they do not give a strong signals over datasets.

\subsection{Relation Types in BATS/Google}
\begin{figure}[h]
    \centering
    \includegraphics[width=\columnwidth]{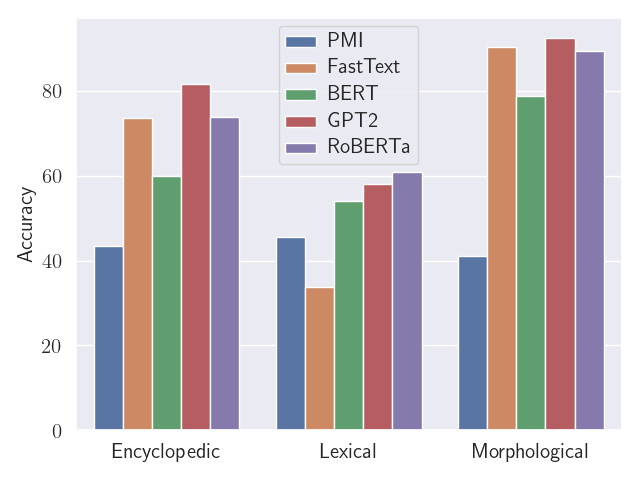}
    \includegraphics[width=\columnwidth]{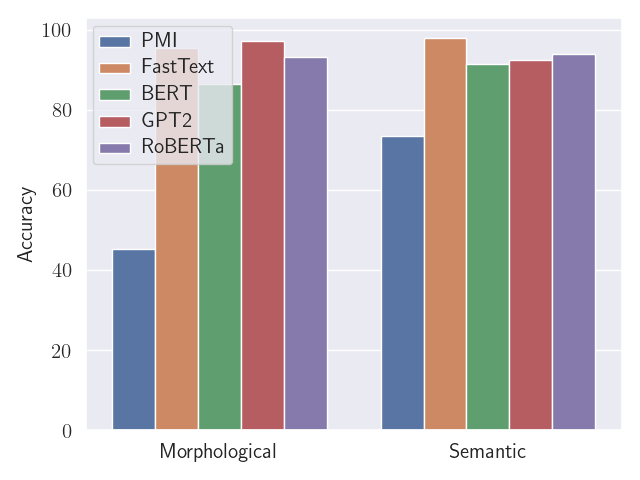}
    \caption{BATS (top) and Google (bottom) results split by high-level categories.}
    \label{fig:bar}
\end{figure}

Figure~\ref{fig:bar} shows the results of different language models with the $s_{\textit{mPPL}}$ scoring function on the different categories of the BATS and Google datasets.

\section{Error Analysis}
Table \ref{tab:example} shows all examples from the U2 dataset of the easiest difficuly (i.e.\ grade 4), which were misclassified by RoBERTa, with $s_{\textit{mPPL}}$ tuned on the validation set. We can see a few typical issues with word embeddings and language models. For instance, in the first example, the model confuses the antonym pair \emph{right:wrong} with synonymy. In the second example, we have that someone who is poor lacks money, while someone who is hungry lacks food. However, the selected candidate pair is \emph{hungy:water} rather than \emph{hungry:food}, which is presumably chosen because water is assumed to be a near-synonym of food. In the third example (\emph{wrench:tool}), the hypnernymy relation is confused with a meronymy relation in the selected candidate \emph{tree:forest}. In the last three examples, the model has selected answers which seem reasonable. In the fourth example, \emph{beautiful:pretty}, \emph{terrible:bad} and \emph{brave:valiant} can all be considered to be synonym pairs. In the fifth example, \emph{vehicle:transport} is clearly the correct answer, but the pair \emph{song:sing} is nonetheless relationally similar to \emph{shield:protect}. In the last example, we can think of being sad as an emotional state, like being sick is a health state, which provides some justification for the predicted answer. On the other hand, the gold answer is based on the argument that someone who is sick lacks health like someone who is scared lacks courage.

\begin{table}[h]
\footnotesize
\begin{tabular}{ll}
\toprule
\textbf{Query}              & \textbf{Candidates}                                                                                                                                                                                                                                       \\
\midrule
hilarious:funny  & \underline{right:wrong}, hard:boring, nice:crazy,     \\
& \textbf{great:good}\\[0.5em]
poor:money       & \textbf{tired:energy}, angry:emotion, hot:ice,                                \\      
&\underline{hungry:water}\\[0.5em]
wrench:tool      & cow:milk, radio:sound, \underline{tree:forest},\\
&\textbf{carrot:vegetable}                                    \\
\midrule
beautiful:pretty & \textbf{terrible:bad}, \underline{brave:valiant}, new:old,   \\
& tall:skinny\\[0.5em]
shield:protect   & computer:talk, \textbf{vehicle:transport},  \\
& pencil:make, \underline{song:sing}\\[0.5em]
sick:health      & \underline{sad:emotion}, tall:intelligence,\\
& \textbf{scared:courage}, smart:energy\\
\bottomrule
\end{tabular}
\caption{Model prediction examples from RoBERTa with $s_{\textit{mPPL}}$ tuned on the validation set. Gold answers are shown in bold, while the model predictions are underlined.}
\label{tab:example}
\end{table}

\end{document}